\documentclass{article}
\usepackage[final,nonatbib]{neurips_2018}

\usepackage[utf8]{inputenc} 
\usepackage[T1]{fontenc}    
\usepackage{hyperref}       
\usepackage{url}            
\usepackage{booktabs}       
\usepackage{amsfonts}       
\usepackage{nicefrac}       
\usepackage{microtype}      
\usepackage{amsmath,amssymb}
\usepackage{algorithmic}
\usepackage{subcaption}
\usepackage{graphicx}

\usepackage{textcomp}
\usepackage{xcolor}
\usepackage{tabularx} 
\usepackage{multirow}

\newcommand{\K}{{\mathcal K}}
\newcommand{\MB}[1]{\mathbf{#1}}
\newcommand{\MBB}[1]{\mathbb{#1}}

\newcommand{\MC}[1]{\mathcal{#1}}

\newcommand{\T}{\top}

\newcommand{\smalleqb}[1]
{
	\begingroup
	\makeatletter
	\def
	\f@size{1}
	{#1}
    \endgroup
}  

\newcommand{\ncl}{c}
\newcommand{\nz}{{\vec{z}}}

\newcommand{\ny}{{\vec{x}}}

\newcommand{\nh}{{\vec{h}}}
\newcommand{\nhs}{h}

\newcommand{\nALF}{\mathbf{B}}
\newcommand{\nAlf}{\vec{\beta}}
\newcommand{\nalf}{{\beta}}
\newcommand{\phih}{\hat{\Phi}}

\title{Large-Margin Multiple Kernel Learning for Discriminative Features Selection\\ and Representation Learning}

\author{%
	Babak Hosseini
	\thanks{
		Preprint of the publication~\cite{hosseini2019large}, as provided by the authors.
		The final publication is available at IEEE Xplore via \url{https://ieeexplore.ieee.org/xpl/conhome.jsp?punumber=1000500}  	
	} \\
	CITEC cluster of excellence\\
	Bielefeld University, Germany\\
	\texttt{bhosseini@techfak.uni-bielefeld.de} \\
	\And
	Barbara Hammer\\
	CITEC cluster of excellence\\
	Bielefeld University, Germany\\
	\texttt{bhammer@techfak.uni-bielefeld.de} \\
}

\begin{document}

\maketitle

\begin{abstract}
Multiple kernel learning (MKL) algorithms combine different base kernels to obtain a more efficient representation in the feature space. 
Focusing on discriminative tasks, MKL has been used successfully for feature selection and finding the significant modalities of the data.
In such applications, 
each base kernel represents one dimension of the data or is derived from one specific descriptor. 
Therefore, MKL finds an optimal weighting scheme for the given kernels to increase the classification accuracy.
%
%
Nevertheless, the majority of the works in this area focus on only binary classification problems or 
aim for linear separation of the classes in the kernel space, which are not realistic assumptions for many real-world problems.
%
In this paper, 
we propose a novel multi-class MKL framework which improves the state-of-the-art by enhancing the local separation of the classes in the feature space.
Besides, by using a sparsity term, 
our large-margin multiple kernel algorithm (LMMK) performs discriminative feature selection by aiming to employ a small subset of the base kernels.  
Based on our empirical evaluations on different real-world datasets,
LMMK provides a competitive classification accuracy compared with the state-of-the-art algorithms in MKL. 
Additionally, it learns a sparse set of non-zero kernel weights which leads to 
a more interpretable feature selection and representation learning.

\textit{Keywords}-Multiple Kernel Learning, Feature Selection, Representation Learning, LMNN.
\end{abstract}

\section{Introduction}	
Multiple kernel learning (MKL) algorithms utilize different data representations in the feature space (base kernels) to obtain an optimal representation upon their combination~\cite{bach2004multiple}. 
We can generally formulate an MKL problem as the minimization of a loss term 
defined in the Reproducing Kernel Hilbert Space (RKHS). 
This cost function usually reflects how separated the data classes are in the RKHS according to a given classification task~\cite{gonen2011multiple}.
Depending on the definition of the problem, MKL can be seen as either finding the 
best parameter values for a specific type of kernel function \cite{jiang2014trace,ye2008multi,niazmardi2017novel,gonen2011multiple}
or learning a weighting vector associated to the pre-computed base kernel~\cite{lin2011multiple,dileep2009representation,xue2017multiple,du2017multiple,xu2009non}.
%

In image processing problems, it is a common practice to derive specific representations by utilizing different types of image descriptors. 
Therefore, an MKL algorithm can learn which descriptors provide more discriminative representations of the data classes~\cite{lin2011multiple,dileep2009representation}. 
Analogously, by computing each base kernel from one specific dimension of the data, MKL can perform discriminative feature selection by assigning larger weights to the most discriminative dimensions of the data~\cite{dileep2009representation,xu2009non,varma2009more,xue2017multiple}.
In practice, any MKL algorithm can also be considered as a multiple kernel feature selection method (MK-FS) provided that it can take pre-computed kernel representations as the inputs.
%
%
%
%

The significant well-studied group of MKL methods is applicable only to the binary-classification problems~\cite{kim2006optimal,aiolli2015easymkl,xue2017multiple,xu2010simple,rakotomamonjy2008simplemkl,dileep2009representation}. 
These algorithms are generally constructed to improve the performance of  
the Support Vector Machines (SVM) as a binary classifier.
%
It is possible to apply these binary MKL methods to multi-class problems throughout defining an ensemble of binary classification tasks, and for each of which train an individual MKL model ~\cite{gu2014model,dileep2009representation,yang2012group}.
However, such strategy results in several kernel combination schemes learned from the individual binary classifiers
and generally does not lead to a unanimous feature embedding.  
%

On the other hand, some recent works have tried to extend MKL to the multi-class problems via defining seamless optimization schemes
by considering all the classes together~\cite{ye2008multi,lin2011multiple,jiang2014trace,gu2015multiple,wang2016discriminative,gu2012representative}.
As a common characteristic, these algorithms try to learn the optimal kernel weights independently of the later on classifier's structure.
Inspired by the Fisher Linear Discriminant Analysis (LDA)~\cite{duda1973pattern}, algorithms similar to DKL~\cite{ye2008multi}, MKL-DR~\cite{lin2011multiple} and MKL-TR~\cite{jiang2014trace} are focused on reducing the intra-class covariances via using the scatter matrices of data in different RKHSs.
In particular, the MKL-DR and MKL-TR methods employ low-dimensional projections, while the latter also applies the convex combination of the base kernels.
As a different approach, RMKL method~\cite{gu2012representative} performs singular value decomposition to find the base kernels which lead to maximum variation in the space spanned by them.
It is claimed that this decomposition finds a more discriminative kernel combination than the original RKHS. 
Similarly, KNMF-MKL~\cite{gu2015multiple} was proposed by reformulating the RMKL approach using the non-negative matrix factorization framework (NMF)~\cite{lee2001algorithms}.

To emphasize the noteworthy shortcomings of the existing MKL algorithms, 
we distinguish them into two general categories: 

%
First, algorithms similar to \cite{dileep2009representation,xue2017multiple,rakotomamonjy2008simplemkl,aiolli2014learning,aiolli2015easymkl} focus on learning a multiple kernel mapping to a target RKHS in which a classifier can linearly separate the different classes from each other.
This objective coincides with the basic principle of the kernelized SVM's structure~\cite{cristianini2000introduction} which is the linear separation of the classes in the feature space.
Nevertheless, obtaining such an ideal representation is usually not affordable for real-world data, or it demands considerable domain knowledge for the specific design of such efficient kernels.
This category generally includes binary MKL algorithms.

%
Second,
another group of MKL methods includes algorithms such as~\cite{kim2006optimal,ye2008multi,lin2011multiple,wang2016discriminative} which follow methodologies analogous to the kernelized LDA's design scheme~\cite{mika2001mathematical}.
They generally try to obtain a multiple kernel representation in a way that the class distributions in RKHS would be generally condensed.
This strategy is effective, especially for multi-class problems.
%
Nevertheless, as a common observation in real data, 
some classes consist of sub-clusters which are located on different regions of the space, 
but are yet well separated from other classes (e.g., having an XOR distribution in the feature space). 
In such cases, it is generally difficult to find a target RKHS in which the classes are globally condensed, especially without doing any feature engineering~\cite{tsang2006efficient}.
This shortcoming is fundamentally problematic for the classifiers which rely on linear separability of the classes (e.g., SVM).
%

By deriving each base kernel from a different source of information in the data, it is highly possible to observe substantial redundancy between these representations\cite{du2017multiple}.
Therefore, it is desirable to reduce this redundancy in favor of the model's interpretation and its discriminability.
In the works similar to SimpleMKL \cite{rakotomamonjy2008simplemkl} and class-specific MKL \cite{liu2016class}, they imposed sparsity on the weights of the base kernels by 
using a convex combination in the MKL problem.
As an improvement, Group Lasso-MKL fused the MKL problem with the $l_p$-norm based on the group Lasso optimization \cite{tibshirani1996regression} to better enforce the sparsity concern \cite{xu2010simple}.
In comparison, SparseRMKL \cite{gu2014model} benefits from an $l_1$-norm constraint in its optimization framework, which provides a better classification performance as well as an enhanced interpretation by specifying the most discriminative contributions among the set of the base kernels.

\subsection{Motivation and Contributions}
Metric learning is the idea of finding an appropriate distance metric 
which transforms the data into a new space in which 
the data distribution provides a more smooth labeling
than the original space
~\cite{chopra2005learning,shental2002adjustment,goldberger2005neighbourhood}.
Based on practical evidence, performing metric learning 
can notably enhance accuracy of distance-based classifiers (e.g., $k$NN) on the test data
even by applying a linear mapping on the input space 
\cite{goldberger2005neighbourhood,shalev2004online}. 
One of the successful distance metric learning algorithms is the Large-Margin Nearest Neighbor (LMNN) which increases the maximum margin between the data instances of different classes~\cite{Weinberger2009}.
In contrast to the global separation of the classes via a hyperplane in SVM, 
LMNN learns a distance metric which improves the local separation of the classes in small neighborhoods of the space.
According to~\cite{Weinberger2009}, the LMNN's resulted metric can improve the $k$NN's classification accuracy even in comparison with the kernelized SVM.
Therefore, we expect that employing metric learning in MKL framework could result in an RKHS in which the $k$NN's discriminative performance can outperform other MKL models.
%
	


\textbf{Contributions:}
In this work, we introduce the metric learning concept to the MKL problem by optimizing a diagonal Mahalanobis metric in the feature space.
Our proposed large-margin multiple kernel algorithm (LMMK) improves the local separation of the classes in a resulted RKHS,
in which it imposes a large margin between data vectors from the different classes.
The specific formulation of LMMK converts the above metric learning problem into finding an optimal combination of the given base kernels in an MKL framework.
It is a multi-class MKL method which results in an efficient data representation for the $k$NN classifier in the feature space. 
%
%
Furthermore, by employing a sparsity term in the convex optimization framework of LMMK,
it behaves as an effective MK-FS algorithm. 
More precisely, it selects the small subset of essential features to enhance the described local class-separation objective.
\section{Preliminaries}
\subsection{Multiple Kernel Learning}
The training set  $\{(\ny_i,\nhs_i)\}_{i=1}^{N}$ 
includes $N$ data samples $\ny_i \in \MBB{R}^n$,
where $\nhs_i \in \{1,2,\dots,\ncl\}$ denotes the corresponding label of $\ny_i$ in a $\ncl$-class setting.
Implicitly, we can assume $d$ non-linear mapping functions 
$\{\Phi_m:\MBB{R} \rightarrow \MBB{R}^{f_m}\}_{m=1}^d$ exist which map $\ny$ into individual RKHSs~\cite{bach2004multiple,wang2014feature}.
Therefore, we can obtain a scaling of the feature space based on the following weighted concatenation:
\begin{equation}
\begin{array}{l}
\phih(\ny)=
[\sqrt{\nalf_1} \Phi_1^\top(\ny),\dots,
\sqrt{\nalf_d} \Phi_{d}^\top(\ny)]^\top,
\end{array}
\label{eq:mk}
\end{equation}
where 
$\phih(\ny)$ 
is the implicit mapping to the resulted RKHS, and $\nAlf$ is the combination vector.
Due to the finiteness of training samples $\ny_i$ the target of each implicit mapping $\Phi_m$ is assumed a finite-dimensional Hilbert space which validates the concatenation of the embeddings in Eq.~(\ref{eq:mk}).
By relating each $\Phi_m(\ny)$ to a kernel function $\K_m(\ny_i,\ny_j)=\Phi_m^\top(\ny_i)\Phi_m(\ny_j)$, 
we can compute the weighted kernel function $\hat{\K}(\ny_i,\ny_j)$ corresponding to $\phih(\ny)$ as the additive combination~\cite{dileep2009representation}
\begin{equation}
\hat{\K}(\ny_i,\ny_j)=\sum_{m=1}^{d} \nalf_m \K_m(\ny_i,\ny_j).
\label{eq:K_alf}
\end{equation}
%

Generally, one can formulate the MKL frameworks as variants of the following optimization
\begin{equation}
\begin{array}{ll}
\nAlf = \underset{\nAlf \in \MB{S}}{\arg\min}& loss(\{\{\K_m(\ny_i,\ny_j)\}_{i,j=1}^N\}_{m=1}^d,\nAlf,\nh),
\end{array}
\label{eq:mk_opt}
\end{equation}
in which the $loss$ term is a cost function that its minimization reflects the given classification task and is also defined by considering the classifier's model. 
The set $\MB{S}$ defines the set of employed constraints on $\nAlf$ based on the MKL algorithm.

If we apply each kernel function $\K_m$ only on the $m$th dimension of the training data (resulting in $n$ feature-kernels), we can assume each corresponding $\Phi_m$ in Eq. (\ref{eq:mk}) maps the $m$th dimension of the data into one individual RKHS.
%
In that case, each solution for Eq. (\ref{eq:mk_opt}) represents a weighted feature selection obtained by the MKL algorithm based on the defined discriminative function $loss$ and the constraints in $\MB{S}$.
It is practical to apply a non-negativity constraint on each $\nalf_m$ to make the resulted kernel weights interpretable as the relative importance of each feature representation to the given discriminative task~\cite{gonen2011multiple}.
Furthermore, including sparsity terms in Eq.~(\ref{eq:mk_opt}) can decrease the redundancy in the above importance profile~\cite{du2017multiple,rakotomamonjy2008simplemkl,liu2016class,xu2010simple,gu2014model}. 
For instance, if the individual RKHSs are correlated, their corresponding entries in $\nAlf$ are preferred to be considerably sparse.

\subsection{Large-Margin Nearest Neighbor}\label{sec:LMNN}
\begin{figure}[t]
	\centering		
	\includegraphics[width=.91\linewidth]{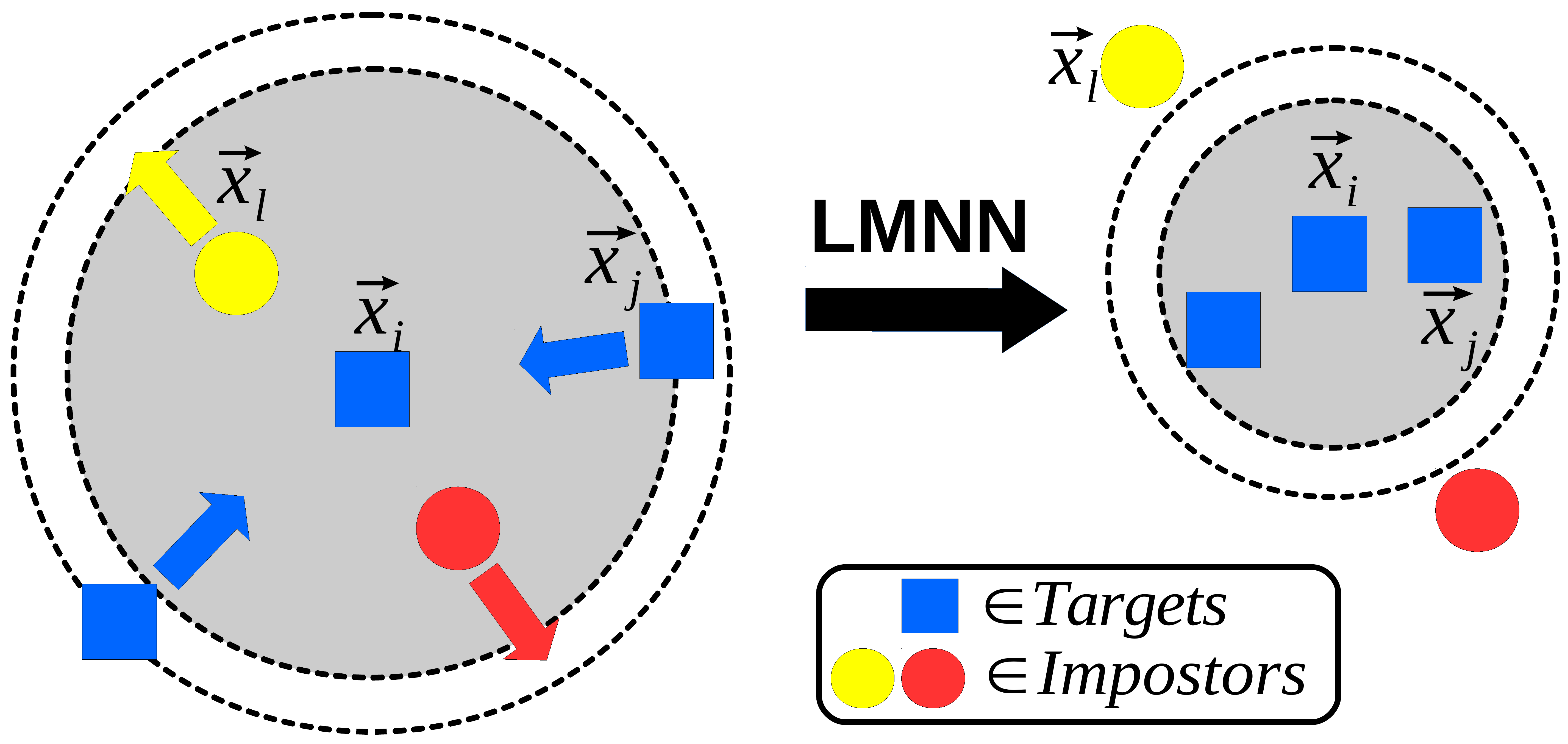}
	\caption{\textbf{Left:} The original distribution of the neighboring points around $\ny_i$, where the rectangles (\textit{targets}) have the same label as that of $\ny_i$, and circles (\textit{impostors}) have different labels.
	\textbf{Right:} The distribution resulted from LMNN's mapping, in which \textit{targets} are closer to $\ny_i$ while \textit{impostors} are pushed farther away from it.
		}
	\label{fig:lmnn}
\end{figure}

The LMNN  algorithm learns the Mahalanobis distance metric 
\begin{equation}
\MC{D}_\MB{L} (\ny_i,\ny_j)= (\ny_i-\ny_j)^\top \MB{L}^\T \MB{L}(\ny_i-\ny_j)
\label{eq:mah}
\end{equation}
throughout finding the linear mapping matrix $\MB{L}\in \MBB{R}^{n \times n}$~\cite{Weinberger2009}. 
For each $\ny_i$, LMNN tries to map it closer to the data samples belonging to the class $\nh_i$ (\textit{targets}),
while pushing it away from the data points with labels other than $\nh_i$
(\textit{impostors}) (Figure~\ref{fig:lmnn}).
%
To that aim, LMNN uses the following convex optimization:
\begin{equation}
\begin{array}{ll}
\underset{\MB{L}}{\min}  &(1-\mu)\underset{i,j \in \MC{N}^{k}_{i}}{\sum} \MC{D}_\MB{L}(\vec x_i,\vec x_j) + 
\mu \underset{i,j \in \MC{N}^{k}_{i}}{\sum}
\underset{l\in\MC{I}^{k}_{i}}{\sum}\xi_{ijl}\\
\mathrm{s.t.} & \MC{D}_\MB{L}(\vec x_i,\vec x_l) - \MC{D}_\MB{L}(\vec x_i,\vec x_j) \ge 1-\xi_{ijl}\\
&\xi_{ijl}\ge 0 ,
\label{eq:lmnn}
\end{array}
\end{equation}
in which $\MC{N}^{k}_{i}$ and $\MC{I}^{k}_{i}$ contain the indices of the $k$-nearest \textit{targets} and \textit{impostors} of $\vec{x}_i$ respectively.
The scalar $\mu\in[0~1]$ makes a trade-off between the pulling (first) and pushing (second) parts of the objective in Eq. (\ref{eq:lmnn}).
Additionally, each positive slack variable $\xi_{ijl}$ is related to a triple $(\ny_i,\ny_j,\ny_l)$, in which $\ny_j$ and $\ny_l$ are respectively a \textit{target} for $\ny_i$ and an \textit{impostor} which is located between $\ny_i$ and $\ny_j$ (similar to Figure~\ref{fig:lmnn}-left).
The scalars $\xi_{ijl}$ model the costs induced by the existing impostors.
\section{Large-Margin Multiple Kernel Learning}
We apply the metric learning concept to the data distribution in the feature space, 
such that it results in having dense neighborhoods of classes 
in which the different classes can be locally separated.
%
Assuming that the dimensions of the feature space are related to individual RKHSs as in Eq. (\ref{eq:mk}), we employ metric learning to find the effective $\nAlf$ that serves the above purpose. 
However, direct application of Eq. (\ref{eq:lmnn}) in the feature space has the following limitations:
%

First, via applying the Mahalanobis metric of Eq.~(\ref{eq:mah}) to the feature space, 
the dimensions of the resulted $\phih(\ny)$ lose their interpretability.
Denoting $\Phi(\ny)$ as the non-weighted concatenation of the base kernels in Eq.~(\ref{eq:mk}) (setting $\nalf_m=1~\forall m$),
%
\begin{equation}
\phih(\ny)(i)=\sum_j l_{ij} \Phi(\ny)(j)
\quad \text{having}~ \phih(\ny)=\MB{L}\Phi(\ny),
\label{eq:mah_prob}
\end{equation}
in which $\Phi(\ny)(i)$ and $\phih(\ny)(i)$ denote the $i$th dimension of $\Phi(\ny)$ and $\phih(\ny)$ respectively in the feature space, 
and $l_{ij}$ indicates the $i$th row from the $j$th column of $\MB{L}$.
%
Consequently, each dimension of $\phih(\ny)$ in the resulted RKHS loses its physical interpretation, 
as it is a weighted combination of the dimensions of the original RKHS.
%

Second, computing Eq. (\ref{eq:mah}) in the feature space (as in Eq. (\ref{eq:mah_prob})) requires explicit access to the dimensions of each $\Phi_i(\ny)$ in the feature space. This requirement cannot be directly fulfilled as it is contrary to our assumption about the implicit definition $\Phi_i(\ny)$.
%

%
To overcome the above issues, we propose the following optimization scheme: 
\begin{equation}
\begin{array}{ll}
\underset{\nAlf}{\min}  &(1-\mu)\underset{i,j \in \MC{N}^{k}_{i}}{\sum} \MC{D}^\phi_{\nAlf}(\ny_i,\ny_j) \\
&+\mu \underset{i,j \in \MC{N}^{k}_{i}}{\sum}
\underset{l\in\MC{I}^{k}_{i}}{\sum}\xi_{ijl}
+\lambda \sum_m \nalf_m
\\
\mathrm{s.t.} & \MC{D}^\phi_{\nAlf}(\vec x_i,\vec x_l) - \MC{D}^\phi_{\nAlf}(\vec x_i,\vec x_j) \ge 1-\xi_{ijl}\\
&\xi_{ijl}\ge 0,~~\nalf_m \ge 0
.
\end{array}
\label{eq:lmmk}
\end{equation}
In Eq. (\ref{eq:lmmk}), the distance metric $\MC{D}^\phi_{\nAlf} (\ny_i,\ny_j)$ is defined in the feature space as:
\begin{equation}
\MC{D}^\phi_{\nAlf} (\ny_i,\ny_j)= [\Phi(\ny_i)-\Phi(\ny_j)]^\top \nALF[\Phi(\ny_i)-\Phi(\ny_j)],
\label{eq:diag_mt}
\end{equation}
where $\MB{\nALF}$ is a diagonal matrix formed based on the entries of $\nAlf$. 
Eq.~(\ref{eq:diag_mt}) defines a Mahalanobis metric in the feature space with a diagonal covariance matrix $\MB{B}$. Therefore, we name $\MC{D}^\phi_{\nAlf} (\ny_i,\ny_j)$ a diagonal metric.
Consequently, each learned $\nalf_m$ in Eq.~(\ref{eq:lmmk}) acts as a selection weight for the $m$th representation of the data in the original RKHS to locally discriminate the classes in the feature space (similar to Figure~\ref{fig:lmnn}).
Additionally, the last objective term in this optimization problem applies an $l_1$-regularization to enforce the selection of the most relevant feature-kernels $\Phi_m(\ny)$ to the defined discriminative objective. 
Therefore, our LMMK framework in Eq.~(\ref{eq:lmmk}) is an MKL optimization problem which is designed for discriminative feature selection and representation learning.
%
\subsection{Optimization}
Based on Eq.~(\ref{eq:K_alf}), the pair-wise distance between each couple of $(\ny_i,\ny_j)$ in the feature space is computed as
\begin{equation}
\begin{array}{l}
\MC{D}^\phi_{\nAlf}(\ny_i,\ny_j)=\\
\sum_{m=1}^{d} \nalf_m [\K_m(\ny_i,\ny_i)+\K_m(\ny_j,\ny_j)-2\K_m(\ny_i,\ny_j)].
\end{array}
\label{eq:D_RKHS}
\end{equation}
Hence, we can compute $\MC{D}^\phi_{\nAlf}(\ny_i,\ny_j)$ without performing any explicit calculation in the feature space in contrast to Eq.~(\ref{eq:mah}).
In addition, by normalizing the kernel matrices of the training set, we have 
$\K_m(\ny_i,\ny_i)=1$ for all the input vectors and base kernels.
Therefore, after eliminating the constant terms, the optimization problem of Eq.~(\ref{eq:lmmk}) is simplified to 
\begin{equation}
\begin{array}{ll}
\underset{\nAlf}{\min}  
&(1-\mu)(\underset{i,j \in \MC{N}^{k}_{i}}
{\sum} [1-\K_{(:)}(\ny_i,\ny_j)])\nAlf \\
&+\mu \underset{i,j \in \MC{N}^{k}_{i}}{\sum}
\underset{l\in\MC{I}^{k}_{i}}{\sum}\xi_{ijl}
+\lambda \sum_{m=1}^d \nalf_m
\\
\mathrm{s.t.} 
& 2[1+\K_{(:)}(\ny_i,\ny_j)-\K_{(:)}(\ny_i,\ny_l)]\nAlf
\ge 1-\xi_{ijl}\\
&\xi_{ijl}\ge 0,~~\nalf_m \ge 0
,
\end{array}
\label{eq:lmmk_lin}
\end{equation}
where 
$\K_{(:)}(\ny_i,\ny_j)=[\K_1(\ny_i,\ny_j),\dots,\K_d(\ny_i,\ny_j)] \in \MBB{R}^d$.
This optimization framework is a convex problem subject to the advance selection of the \textit{targets} and \textit{impostors} which are 
indexed by $\MC{N}^{k}_{i}$ and $\MC{I}^{k}_{i}$ respectively.
Hence, it is an instance of the non-negative linear programming (LP),
and we can efficiently optimize it via using solvers such as YALMIP~\cite{lofbergtoolbox} or CVX~\cite{grant2008cvx}.
Additionally, similar to a practical hint from \cite{Weinberger2009},
we repeat the optimization loop for a few iterations while updating $\MC{N}^{k}_{i}$ and $\MC{I}^{k}_{i}$ at the end of each run. 
These few extra repetitions can lead to more optimal solutions.
For the efficient implementation of Eq.~(\ref{eq:lmmk_lin}), the code of LMMK algorithm would be accessible via an online public repository\footnote{https://github.com/bab-git/LMMK}.
%
\subsection{Classification of Test Data}
We perform the classification of each test data sample $\nz$ by using the $k$NN algorithm based on the distances in the resulted RKHS. 
To that aim, we compute $\MC{D}^\phi_{\nAlf}(\nz,\ny_i)$ as the distance between $\nz$ and each training sample using the learned diagonal matrix $\nALF$ in the feature space analogous to Eq.~(\ref{eq:D_RKHS}).
\subsection{Complexity and Convergence of LMMK}
The optimization framework of Eq.~(\ref{eq:lmmk_lin}) is an LP problem, and consequently, it converges in limited $t$ steps to an optimal solution. 
On the other hand, an LP solver optimizes $\nAlf$ with the 
computational complexity of $\MC{O}(t(2d+3N_l)+dN_j+2dN_l)$, in which $N_l$ and $N_j$ are the total number of \textit{targets} and the size of $\vec\xi$ respectively. 
Based on the definition of the \textit{targets} and \textit{impostors}, 
we have $N_l \approx \frac{N^2(\ncl-1)}{\ncl}$ and $N_j=kN$. 
In addition, for common real-world datasets we observe $N>>t$ in practice;
hence, the total time complexity of the algorithm is approximately $\MC{O}(N^2)$.
This complexity is almost comparable to that of computing the base kernel matrices for each dataset before running the algorithm.
\section{Experiments}
In this section, we implement our proposed LMMK algorithm 
on different real-world datasets and evaluate its performance by carrying out empirical comparisons to other MKL alternative algorithms.
To that aim, we consider two different scenarios for our experiments: 
\begin{enumerate}
	\item Representation learning, in which we compute the base kernels upon different types of image descriptors on each the dataset.
	Hence, the results of MKL frameworks are interpreted as the most discriminative descriptors they select for each dataset. 
	\item Feature selection, where each base kernel is computed using one specific dimension of the data ($d=n$), and MKL methods are expected to assign larger weights to the more discriminative features of the data.
\end{enumerate}
In both scenarios, all the base kernels are computed using the Gaussian kernel function
 \begin{equation}
\K_m(\ny_i,\ny_j)=exp(-\MC{D}(\ny_i^m,\ny_j^m)^2/\delta_m),
 \label{eq:gausker}
 \end{equation}
in which
$\mathcal{D}(\ny^m_i,\ny^m_j)$ indicates the pairwise distance between
$(\ny_i,\ny_j)$ based on the $m$th representation of the input data,
and $\delta_m$ denotes the average of $\MC{D}(\ny_i^m,\ny_j^m)$ for all data samples. 

\subsection{Datasets}
According to the discussed implementation scenarios, we choose two different types of datasets: 1) Image datasets for representation learning, 2) Multidimensional time-series (MTS) for discriminative feature selection.

Regarding image datasets, we make the following selection:
\begin{itemize}
	\item Caltech-101~\cite{fei2007learning} is a collection of 101 object categories which includes 40 to 800 images per class of object.
	The high inter-class variations within this dataset make it a challenging image classification benchmark.	
	For our experiments, we choose 5 different training subsets with the sizes of 5, 10, 15, 20 25, and 30 images per class, and a testing subset of 15 images per category.
	\item Pascal VOC 2007~\cite{everingham2006pascal} is a dataset consisting of 20 different classes of objects and is related to a classification challenge. Out of 9,963 imaged, we employ 50$\%$ of the samples for training and the rest for testing as provided in~\cite{everingham2006pascal}.
	\item Oxford Flowers17~\cite{nilsback2006visual} is a collection of images related to 16 different species of flowers and are composed of 80 images per category.
	The large intra-class variations for some flower species causes substantial overlapping instances in this dataset.
	As a common practice in the literature~\cite{nilsback2006visual}, we select 40 pre-defined images per class for training and preserve the rest for testing.
\end{itemize}	

To evaluate LMMK's performance for the discriminative feature selection scenario, we select the following real-world MTS datasets:
\begin{itemize}
	\item PEMS dataset~\cite{cuturi2011fast} consists of the daily traffic information related to San Francisco bay freeways, and the classification task is to determine the correct day of the week related to each data sequence. It has 963 dimensions and 60 sequences per each of the 7 class.
	\item AUSLAN is an MTS dataset from the UCI repository~\cite{Dua:2017} containing 95 classes of Australian language signs. It includes 2565 samples of 128-dimension MTS sequences.
	\item UTKinect is a dataset of human action recognition~\cite{xia2012view} including 60-dimension Kinect-based skeleton sequences related to 10 different actions, where each class contains 20 MTS sequences.		
\end{itemize}	
\subsection{Baseline Algorithms}	
To have a proper evaluation of our proposed method, we make our comparison between LMMK and the following major MKL algorithms:
MKL-TR~\cite{jiang2014trace},
MKL-DR~\cite{lin2011multiple},
DMKL~\cite{wang2016discriminative},
KNMF-MKL~\cite{gu2015multiple}, and
RMKL~\cite{gu2012representative}.
These algorithms are designed for multi-class MKL problems; hence, we can inspect their results from feature selection and representation learning perspectives.
Also, as the baseline classifiers, we implement multi-class SVM~\cite{chang2011libsvm} and $k$NN using the average of the base kernels resulting in SVM-ave and $k$NN-ave respectively.
\\\textbf{Note:} Although there exist various deep learning classifiers or object detection methods specially designed for image datasets, they do not fit the multiple kernel scope of our comparisons. 
Nevertheless, as a suggested extended experimental setting, one can use those methods as rich feature extraction techniques to obtain more discriminative base kernels for the MKL methods.

\subsection{Experimental Setup}	
We evaluate the performance of the selected MKL algorithms based on classification accuracy 
$Acc=\frac{\#\text{correct predictions}}{\# \text{all data samples}}$
by taking the average of 10 random repetitions for each dataset.
The LMMK algorithm's hyper-parameters ($k,\mu,\lambda$) are tunned throughout performing cross-validation (CV) on the training set.
However, based on practical evidence (Sec.~\ref{sec:par_sens}), having $0.5 \le \mu \le 0.7$ and tuning $1\le k \le 5$ can lead to satisfactory performance.
Furthermore, we advise the reader to tune ($\mu,k$) first and find the optimal sparsity weight ($\lambda$) afterward. The above strategy can significantly reduce the parameter search space.
Likewise, we tune the hyper-parameters of the baselines based on CV on the training set. 
\subsection{Representation Learning}
We perform our representation learning experiments on the selected image datasets, for which the base kernels are computed upon a set of image descriptors.
To that aim, the distance $\MC{D}(\ny_i^m,\ny_j^m)$ in Eq.~(\ref{eq:gausker}) is computed as the Euclidean distance between $(\ny_i,\ny_j)$ after applying the $m$th descriptor to the data.

\subsubsection{Caltech-101}
For the Caltech-101 dataset, we adopt the following 10 different image descriptors with specifications explained in~\cite{lin2011multiple}: 
SIFT-Dist~\cite{lowe2004distinctive},
SIFT-SPM~\cite{lazebnik2006beyond},
PHOG~\cite{bosch2007image}, 
C2-SWP~\cite{serre2006object},
C2-ML~\cite{mutch2006multiclass},
GB-Dist~\cite{berg2001geometric},
GB,
SS-Dist/SS-SPM~\cite{shechtman2007matching},
and GIST~\cite{oliva2001modeling}. 
Table~\ref{tab:cal_ac} reports the accuracies of the MKL methods for the Caltech-101 dataset. In addition to the multi-class MKL methods, we also included the accuracy rates for some of the published binary MKL techniques for this dataset such as
SimpleMKL~\cite{rakotomamonjy2008simplemkl},
Lasso-MKL~\cite{xu2010simple},
and GS-MKL~\cite{yang2012group}.
Based on the results, LMMK algorithm outperforms all other baselines on the majority of the experiments. Its performance is $4\%$ higher than the best method (MKL-TR) when 30 training samples are used per class ($N_{tr}=30$). 
Table~\ref{tab:cal_ac} shows that the focus of LMMK on local separation of the classes was effective against the existing large intra-class variations in the Caltech-101 dataset.
However, LMMK's performance becomes comparable or slightly lower than the best methods when the per class training samples are sparse.
In those cases, the neighborhood distributions do not coincide with the class labeling anymore, which is not a proper training condition for the algorithms relying on $k$NN predictions.

Table~\ref{tab:cal_d} shows the normalized kernel weights assigned to each descriptor after implementation of LMMK on the Caltech-101 dataset ($N_{tr}=30$). 
Besides, it includes the $k$NN accuracies when using each base kernel individually, which approximately reveal the weak and strong descriptors for this dataset. 
Based on this table, LMMK generally assigned larger weights to the more discriminative descriptors (e.g., GB-DIS and SIFT-DIS). 
Additionally, its sparsity term eliminates the use of weak kernels (e.g., C2-SWP and PHOG) and also reduces the possible discriminative redundancies among the strong descriptors (e.g., GB and SIFT-SPM).
However, our MKL algorithm still keeps GIST descriptor despite its mediocre quality. 
Therefore, we conclude that this descriptor provides an effective complement to other selected base kernels concerning local separation of the classes in the RKHS.

\begin{table}
	\centering
	\caption{Caltech-101: Comparison of classification accuracies ($\%$).}	
		\begin{tabular}{|l|c|c|c|c|c|c|} %
		\hline
		\multirow{2}{*}{Method}	& \multicolumn{6}{c|}{$\#$Training samples per class ($N_{tr}$)}\\
		\cline{2-7}
		& 5 &10&15&20&25&30\\
		\hline
		\hline
		$k$NN-ave&46.1&57.3&64.7&68.2&73.5&76.8\\
		\hline
		SVM-ave&49.7&59.2&64.8&69.7&74.4&77.3\\
		\hline
		DLK(2008)~\cite{ye2008multi}&53.7&62.1&68.2&71.1&74.6&77.9\\
		\hline
		SimpleMKL(2008)~\cite{rakotomamonjy2008simplemkl}&--&53.6&--&63.4&--&76.4\\
		\hline		
		Lasso-MKL(2010)~\cite{xu2010simple}&--&60.1&--&70.7&--&80.7\\
		\hline
		RMKL(2012)~\cite{gu2012representative}&54.7&66.4&71.3&74.3&76.8&78.8\\
	 	\hline		
		KNMF-MKL(2015)~\cite{gu2015multiple}&53.5&65.2&71.5&78.6&79.8&81.1\\		
		\hline
		GS-MKL(2012)~\cite{yang2012group}&--&66.2&75.1&81.5&83.7&84.3\\
		\hline
		MKL-DR(2011)~\cite{lin2011multiple}&58.4&68.8&74.5&77.5&79.8&81.4\\		
		\hline
		DMKL(2016)~\cite{wang2016discriminative}&59.1&\textbf{69.3}&75.2&81.4&83.5&83.7\\
		\hline		
		MKL-TR(2014)~\cite{jiang2014trace}&\textbf{59.8}&\textbf{69.4}&75.8&82.3&84.1&84.6\\
		\hline
		LMMK(\textbf{proposed})&57.6&68.2&\textbf{76.2}&\textbf{84.4}&\textbf{86.2}&\textbf{88.6}\\
		\hline		
	\end{tabular}
	\\ \scriptsize The best results (\textbf{bold}) are according to a two-sample t-test at a $5\%$ significance level.
	\label{tab:cal_ac} 
\end{table}

\begin{table}
	\centering
	\caption{Caltech-101: Normalized kernel weights that LMMK assigned to each image descriptor, and $k$NN accuracy for each base kernel.}	
	\begin{tabular}{|l|c|c||l|c|c|} %
		\hline
		Descriptor&\textit{Acc}&$\nalf$&Descriptor&\textit{Acc}&$\nalf$\\
		\hline
		\hline
		SIFT-Dist&66.9&0.73&GB-Dis&72.3&1.00\\
		\hline
		SIFT-SPM &62.3&0&	GB&67.4&0\\		
		\hline
		PHOG	 &47.5&0&	SS-Dist&64.7&0.15\\		
		\hline
		C2-SWP	 &39.5&0&	SS-SPM&62.3&0\\		
		\hline
		C2-ML	 &57.7&0&	GIST&59.3&0.31\\		
		\hline		
	\end{tabular}
	\label{tab:cal_d} 
\end{table}
\subsubsection{Pascal VOC 2007} 
As the descriptors for Pascal VOC 2007 dataset, we employ 
PHOG~\cite{bosch2007image},
DCSIFT/DSIFT~\cite{lazebnik2006beyond},
SS-Dist~\cite{shechtman2007matching},
and texture feature (Gabor feature~\cite{bau2010hyperspectral}).
In Table~\ref{tab:pas_ac}, the comparison of the classification accuracies on this dataset is provided, which also includes the published results of two binary MKL algorithms Canonical MKL~\cite{bach2004multiple} and GS-MKL~\cite{yang2012group}.
For the Pascal dataset, the LMMK algorithm has a superior performance compared to the MKL baselines. This difference shows that the classes can be better discriminated locally compared to the global discrimination strategies used in other MKL methods.
More precisely, LMMK shows $2.7\%$ and $14.6\%$ increase in accuracy compared to the best method (DMKL) and the $k$NN-ave classifier.

\subsubsection{Oxford Flowers17}
We apply the following 6 descriptors for the Oxford Flowers17 dataset:
DCSIFT~\cite{lazebnik2006beyond},
texture feature~\cite{bau2010hyperspectral},
SS-Dist~\cite{shechtman2007matching},
HOG~\cite{dalal2005histograms},
SIFT-Dist~\cite{lowe2004distinctive},
and HSV color histogram.
Based on the reported results in Table~\ref{tab:pas_ac}, 
both SVM-ave and $k$NN-ave classifiers achieved similar performances using the original RKHS, while using LMMK method boosts $k$NN performance to $93.8\%$ with a margin of $4.3\%$ compared to the best approach (MKL-TR).
This observation implies that the intra-class variations have 
become much smaller in the RKHS resulted from LMMK compared to the original RKHS.
\begin{table}
	\centering
	\caption{Comparison of classification accuracies ($\%$) on Pascal VOC 2007 and Oxford Flowers17 datasets.}	
	\begin{tabular}{|l|c|c|} %
		\hline
		Method	& Pascal VOC& Flowers17\\
		\hline
		\hline
		$k$NN-ave&55.2&81.9\\
		\hline
		SVM-ave&52.6&82.4\\
		\hline
		Can-MKL(2004)~\cite{bach2004multiple}&54.5&--\\
		\hline
		DLK(2008)~\cite{ye2008multi}&56.3&83.5\\		
		\hline
		RMKL(2012)~\cite{gu2012representative}&59.3&85.9\\
		\hline		
		KNMF-MKL(2015)~\cite{gu2015multiple}&61.1&84.6\\
		\hline
		GS-MKL(2012)~\cite{yang2012group}&62.5&--\\
		\hline
		MKL-DR(2011)~\cite{lin2011multiple}&62.5&85.7\\
		\hline
		DMKL(2016)~\cite{wang2016discriminative}&64.7&88.3\\
		\hline		
		MKL-TR(2014)~\cite{jiang2014trace}&64.2&89.5\\
		\hline
		LMMK(\textbf{proposed})&\textbf{69.4}&\textbf{93.8}\\
		\hline		
	\end{tabular}
	\\	\scriptsize The best results (\textbf{bold}) are according to a two-sample t-test at a $5\%$ significance level.
	\label{tab:pas_ac} 
\end{table}

\subsection{Feature Selection}
In our second experimental scenario, we perform discriminative feature selection for MTS datasets using the selected MKL algorithms. 
To that purpose, each Gaussian feature-kernel $\K_m(\ny_i,\ny_j)$ 
%
is computed upon the application of the global alignment kernel~\cite{cuturi2007kernel} on 
the $m$th dimension of the input.
\\\textbf{Note:} There exist state-of-the-art algorithms specifically designed for the classification of MTS. They generally perform temporal segmentations or frame-based analysis of the data samples. Therefore, these algorithms do not belong to the intended multiple kernel scope of our experiments.
In order to evaluate the feature selection performance of the selected baselines, besides the classification accuracy ($Acc$), we also measure the number of selected features of the data (base kernels) via $\|\nAlf\|_0$. 
Consequently, a large $Acc$ along with a small $\|\nAlf\|_0$ describes an
ideal discriminative feature selection, in which the classes could be distinguished with high accuracy while using a few selected features.

Table~\ref{tab:fs} contains the implementation results of the MKL algorithms on the selected MTS benchmarks. The LMMK algorithm outperforms other MKL baselines regarding the classification accuracy. 
It leads to a $4.2\%$ increase in the value of $Acc$ for the UTKinect dataset while this margin is $1\%$ for the AUSLAN dataset. This observation shows that the local class-separation strategy is more effective against the data distribution in the first dataset. 
Also, it significantly increases the performance of $k$NN method especially for the PEM dataset, in which $k$NN-ave has a relatively low accuracy due to its large number of features (963). Nevertheless, LMMK optimization leads to a $15.7\%$ increase in the performance of $k$NN for this dataset. 
Considering other baselines, DMKL and MKL-TR alternatively take the second position in classification accuracy, which shows that the discriminative effect of the low-rank model in MKL-TR may vary depending on the given dataset.

Regarding the feature selection performance, the value of $\|\nAlf\|_0$ has ranked LMMK among the low-feature group of methods (DMKL, MKL-TR, LMMK), which is due to the direct application of an $l_1$-norm sparsity term in the optimization scheme of Eq.~(\ref{eq:lmmk}).
In comparison, DMKL and MKL-TR obtained smaller values for $\|\nAlf\|_0$ in PEM and AUSLAN datasets respectively, but they showed lower $Acc$ in return. 
Therefore, we can claim that LMMK achieves more discriminative feature-selections even for these cases.
%
To explain the feature selection results of other baselines, DMKL and MKL-TR use a convex combination constraint on $\nAlf$ which directly enforces sparsity, while MKL-DR and DKL have quadratic constraints on the kernel weights which applies a weaker restriction on the number of non-zero kernel weights.
On the other hand, KNMF-MKL and RMKL do not have any constraint in their optimization framework related to the sparseness of the selected features.
\begin{table}
	\centering
	\caption{Comparison of accuracies ($Acc$) and $\|\nAlf\|_0$ on the MTS datasets.}	
	\begin{tabular}{|l|c|c |c|c| c|c|} %
		\hline				
		\multirow{2}{*}{Method}
		&\multicolumn{2}{c|} {{PEM}} 
		&\multicolumn{2}{c|} {{AUSLAN} } 
		&\multicolumn{2}{c|} {UTKinect}\\
		\cline{2-7}
		& $Acc$&  $\|\nAlf\|_0$ & $Acc$&  $\|\nAlf\|_0$ & $Acc$&  $\|\nAlf\|_0$ \\
		\hline
  	    \hline				
		$k$NN-ave&75.6&963&83.1&128&83.7&60\\
		\hline
		SVM-ave&83.2&963&87.2&128&85.4&60\\
		\hline
		DLK~\cite{ye2008multi}&84.1&171&87.9&79&86.3&41\\	
		\hline
		RMKL~\cite{gu2012representative}&84.9&690&88.7&95&88.3&55\\
		\hline		
		KNMF-MKL~\cite{gu2015multiple}&85.7&742&88.3&101&87.5&52\\
		\hline
		MKL-DR~\cite{lin2011multiple}&86.4&220&89.6&65&88.7&37\\
		\hline
		DMKL~\cite{wang2016discriminative}&88.2&\textbf{64}&91.3&47&90.7&28\\
		\hline		
		MKL-TR~\cite{jiang2014trace}&88.5&81&91.1&\textbf{31}&91.4&25\\
		\hline
		LMMK(\textbf{proposed})&\textbf{91.3}&75&\textbf{92.1}&39&\textbf{95.6}&\textbf{20}\\
		\hline		
	\end{tabular}	
	\\ \scriptsize The best result (\textbf{bold}) is according to a two-sample t-test at a $5\%$ significance level.
	\label{tab:fs} 
\end{table}
\subsection{Effect of Hyper-parameters}\label{sec:par_sens}
\begin{figure}[tb]
	\begin{subfigure}{0.23\textwidth}
		\centering		
		\includegraphics[width=.951\linewidth]{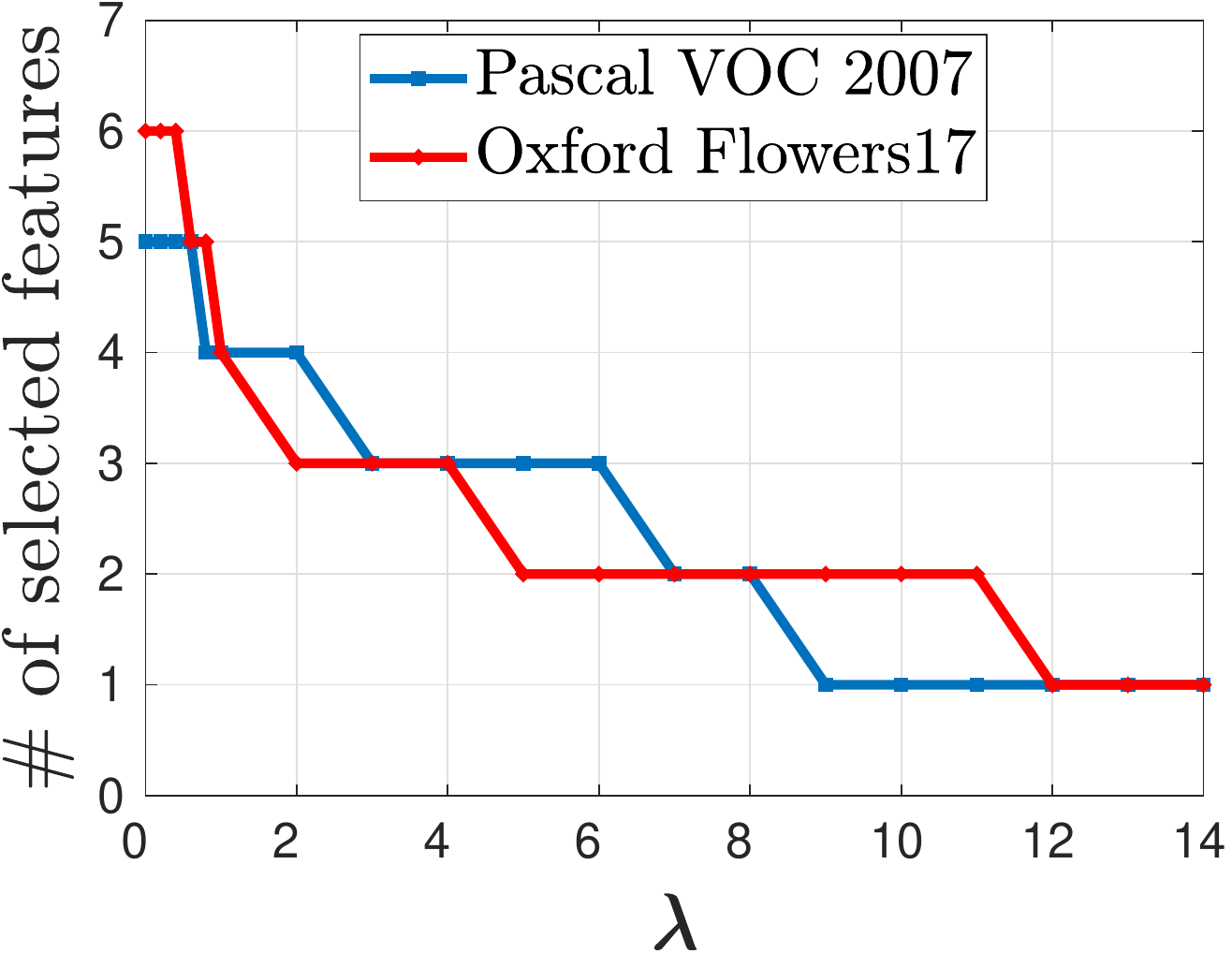}  
		\caption{}
	\end{subfigure}	
	\begin{subfigure}{0.23\textwidth}
		\centering		
		\includegraphics[width=.951\linewidth]{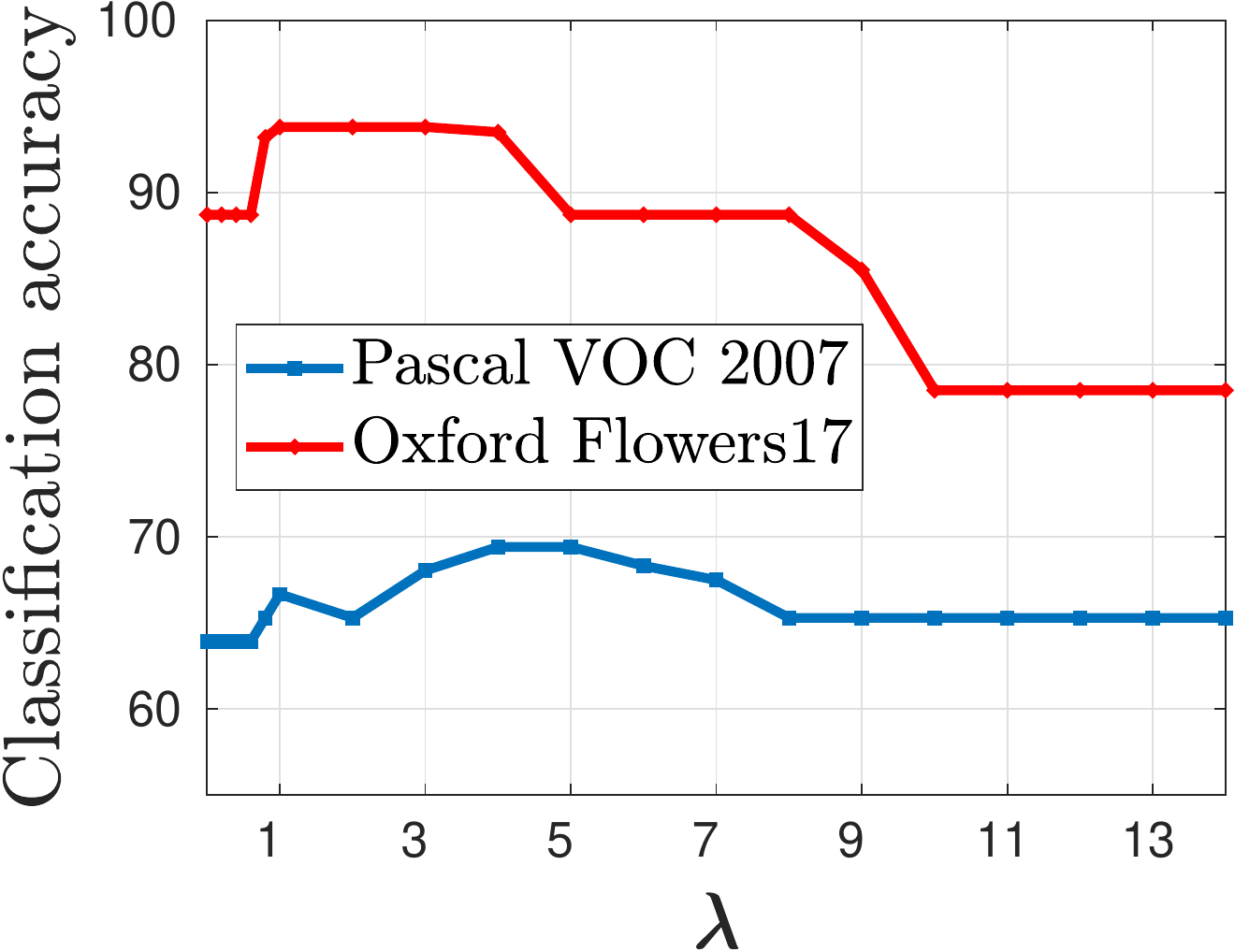}
		\caption{}			
	\end{subfigure}
	\begin{subfigure}{0.23\textwidth}
		\centering		
		\includegraphics[width=.951\linewidth]{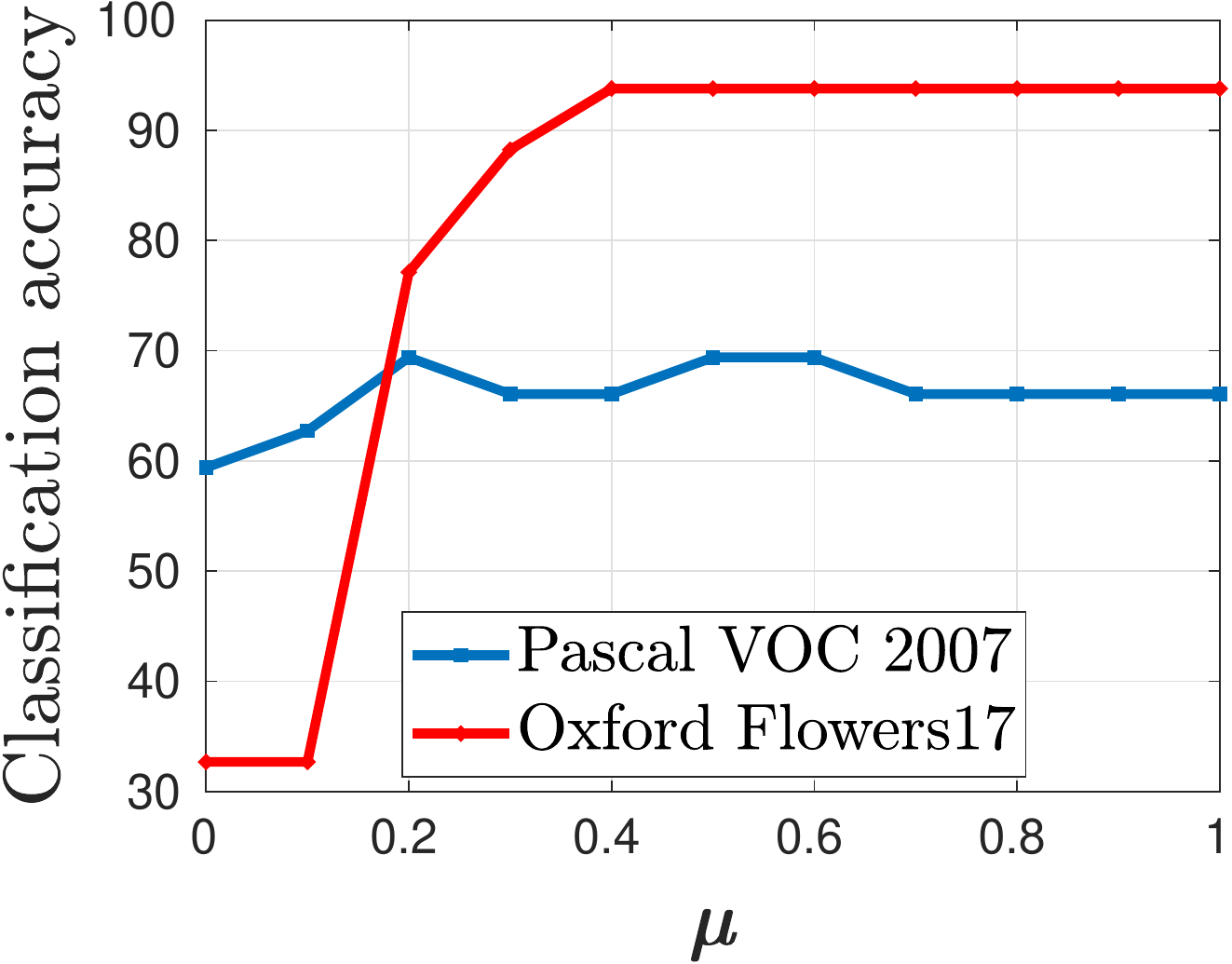}
			\caption{}			
	\end{subfigure}
	\hfill
	\begin{subfigure}{0.23\textwidth}
		\centering		
		\includegraphics[width=.951\linewidth]{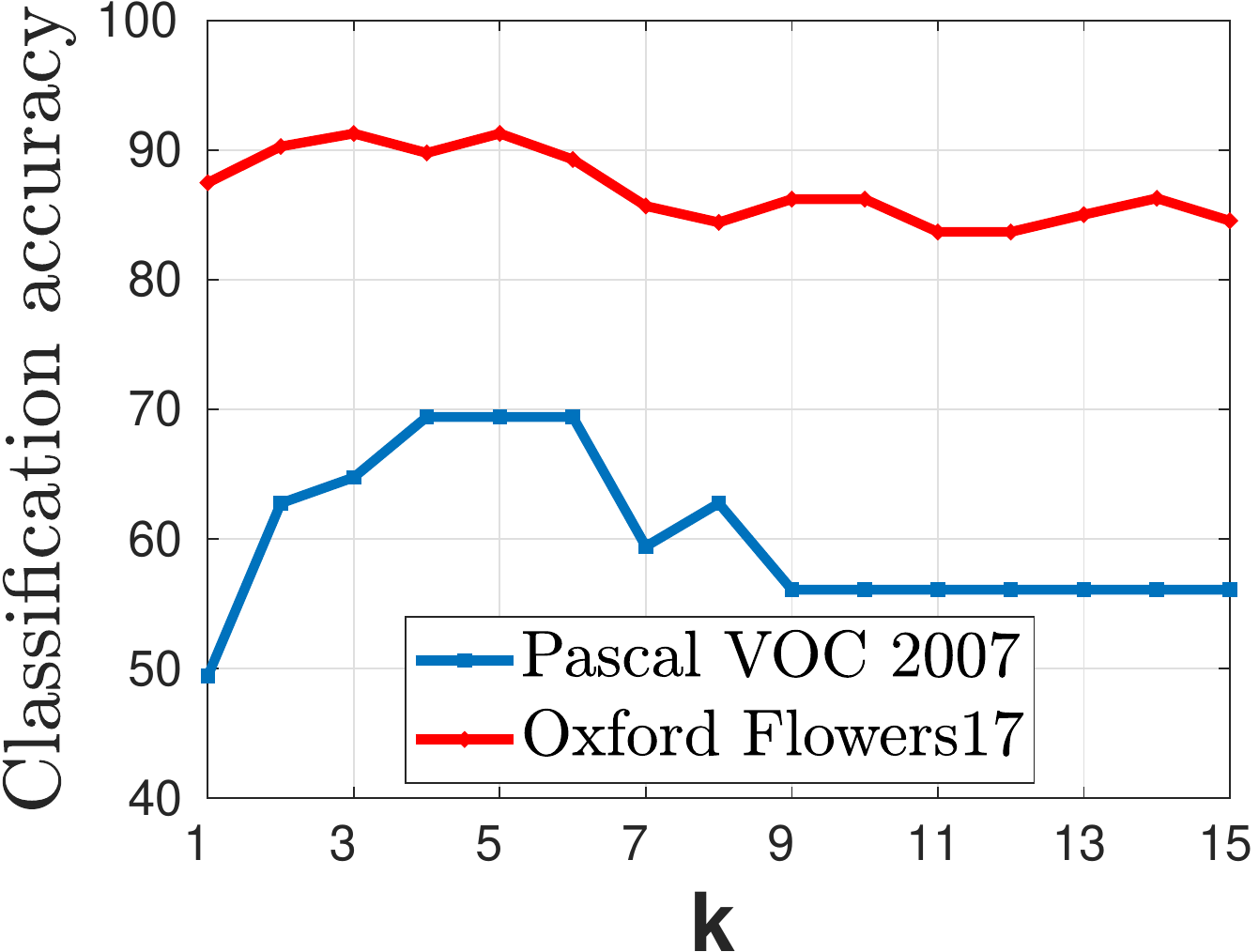}
		\caption{}			
	\end{subfigure}	
	\caption{Effects of parameter changes on LMMK's performance.}		
	\label{fig:sens}
\end{figure}

In this section, we study the effect of the parameters ($\lambda,k,\mu$) on the performance of LMMK.
As described in Figure~\ref{fig:sens}, we perform three experiments on Flowers17 and Pascal datasets, for each of which we study the algorithm's performance by changing one of the above parameters while fixing the two others.

At first, we change $\lambda$ in the range $[0~~14]$ as in Figure~\ref{fig:sens}-a. 
Based on the observations, we conclude that increasing the value of $\lambda$ leads to a stronger sparsity force in Eq.~(\ref{eq:lmmk}) and consequently results in a smaller set of selected features for both of the datasets.
Figure~\ref{fig:sens}-b shows that limited increases in $\lambda$ can improve the classification accuracies, but large values of $\lambda$ would damage the discriminative property of the resulted RKHS.
It is essential to indicate that the points $\lambda=0$ in Figure~\ref{fig:sens}-a and Figure~\ref{fig:sens}-b are related to the performance of LMMK$_{\lambda=0}$, which is the LMMK's algorithm without having the sparsity term in Eq.~(\ref{eq:lmmk}).
Based on the figures, LMMK$_{\lambda=0}$ has the accuracies of $88.7\%$ and $63.9\%$ for Oxford and Pascal datasets, which are comparable to the performances of DMKL and MKL-TR (as the best baselines in Table~\ref{tab:pas_ac}).
This evidence proves our claim regarding the effectiveness of focusing on local discrimination of the classes in the feature space even without the sparsity objective.
%
Additionally, making a comparison between LMMK$_{\lambda=0}$ and sparse LMMK reveals the notable benefit of the $l_1$-norm sparsity term to both feature selection and classification accuracy.

Figure~\ref{fig:sens}-c demonstrates the effect of the trade-off between the first two objective terms in Eq.~(\ref{eq:lmmk}).
For the Pascal dataset, having a balance between the pulling and pushing terms (with $0.5\le \mu \le 0.6$) leads to the highest accuracy.
However, for Flowers17, pushing the impostors away performs a more significant role in local discrimination of the classes (check for $0.5\le \mu \le 1$).  
Based on the experimental observations like the above, tuning $\mu$ around $0.5$ generally results in a good performance.

Based on the classification accuracy curves of Figure~\ref{fig:sens}-d, the best choice for the value of $k$ depends on the distribution of the classes; 
nevertheless, selecting large values for this parameter (e.g., $10\le k$) is expected to reduce the $Acc$.
As the explanation, 
by increasing the size of neighborhoods ($k$), they cannot preserve their local property anymore.
\section{Conclusion}
In this work, we proposed a new multiple kernel algorithm to perform discriminative MKL for
the multi-class problems.
%
%
Our LMMK algorithm focuses on 
improving 
the local separation of the classes in the 
feature space.
To that aim, 
we applied metric learning to
the feature space by defining a diagonal multiple kernel metric in the RKHS.
LMMK finds an efficient weighted combination of the base kernels using an LP optimization framework.
Furthermore, we employed an $l_1$-norm sparsity term in the formulation of LMMK to enforce the compactness 
in choosing the discriminative based kernels.
We implemented our algorithm on the real-world multi-class benchmarks of images and multidimensional time-series. 
The evaluation results show that LMMK outperforms other MKL algorithms regarding representation learning and discriminative feature selection.
\section*{Acknowledgement}
This research was supported by the Cluster of Excellence Cognitive 
Interaction Technology 'CITEC' (EXC 277) at Bielefeld University, which
is funded by the German Research Foundation (DFG).

\bibliographystyle{unsrt}
\bibliography{/vol/semanticma/Thesis/Publications/Ref4Papers_CS}

\end{document}